\documentclass[conference,a4paper]{IEEEtran}
\IEEEoverridecommandlockouts
\usepackage[cmex10]{amsmath}
\usepackage{amssymb,amsfonts}
\usepackage{dblfloatfix}
\usepackage[table,xcdraw]{xcolor}
\usepackage[ruled,vlined]{algorithm2e}
\usepackage{graphicx}
\graphicspath{{Figures/PDF/}{Figures/PNG/}}
\usepackage{array}
\usepackage{booktabs}
\usepackage{siunitx}
\usepackage[numbers,compress]{natbib}
\usepackage{texnames}
\usepackage{bm,bbm}
\usepackage{orcidlink}
\usepackage{tabularray}
\usepackage{multirow}
\usepackage{booktabs}

\begin{document}

\title{Evaluating the Efficacy of Sentinel-2 versus Aerial Imagery in Serrated Tussock Classification}
\author{\IEEEauthorblockN{
        Rezwana Sultana\orcidlink{0009-0006-8464-0698}\textsuperscript{1,2}, 
        Manzur Murshed\orcidlink{0000-0001-7079-9717}\textsuperscript{1}, 
        Kathryn Sheffield\orcidlink{0000-0003-2624-9739}\textsuperscript{2}, 
        Singarayer Florentine\orcidlink{0000-0002-5734-3421}\textsuperscript{3},
        Tsz-Kwan Lee\orcidlink{0000-0003-4176-2215}\textsuperscript{1}, and \\ 
        Shyh Wei Teng\orcidlink{0000-0003-0347-3797}\textsuperscript{3}
    }
\IEEEauthorblockA{\textsuperscript{1}School of Information Technology, 
Deakin University, Australia}
\IEEEauthorblockA{\textsuperscript{2}Agriculture Victoria Research, Department of Energy, Environment and Climate Action, Australia}
\IEEEauthorblockA{\textsuperscript{3}Institute of Innovation, Science and Sustainability, Federation University, Australia}}
\maketitle
\begin{abstract}
Invasive species pose major global threats to ecosystems and agriculture. Serrated tussock (\textit{Nassella trichotoma}) is a highly competitive invasive grass species that disrupts native grasslands, reduces pasture productivity, and increases land management costs. In Victoria, Australia, it presents a major challenge due to its aggressive spread and ecological impact. While current ground surveys and subsequent management practices are effective at small scales, they are not feasible for landscape-scale monitoring. Although aerial imagery offers high spatial resolution suitable for detailed classification, its high cost limits scalability. Satellite-based remote sensing provides a more cost-effective and scalable alternative, though often with lower spatial resolution. This study evaluates whether multi-temporal Sentinel-2 imagery, despite its lower spatial resolution, can provide a comparable and cost-effective alternative for landscape-scale monitoring of serrated tussock by leveraging its higher spectral resolution and seasonal phenological information. A total of eleven models have been developed using various combinations of spectral bands, texture features, vegetation indices, and seasonal data. Using a random forest classifier, the best-performing Sentinel-2 model (M76*) has achieved an Overall Accuracy (OA) of 68\% and an Overall Kappa (OK) of 0.55, slightly outperforming the best-performing aerial imaging model's OA of 67\% and OK of 0.52 on the same dataset. These findings highlight the potential of multi-seasonal feature-enhanced satellite-based models for scalable invasive species classification.
\end{abstract}


\section{Introduction}
Serrated tussock (\textit{Nassella trichotoma}), classified as a Weed of National Significance by the Australian Government, is an aggressive invasive grass species known for its ability to outcompete native vegetation, reduce pasture productivity, and alter ecological balance\cite{osmond2008serrated, humphries_comparative_2021}. In regions like Victoria, Australia, the spread of serrated tussock poses a significant threat to biodiversity and productive land systems. Effective control requires accurate and timely monitoring at landscape scale level. However, a key challenge in detecting serrated tussock lies in is its spectral and structural similarity to native tussock grass species \cite{VSTWP2023}. 

Traditional ground-based surveys, while accurate, are labour-intensive and financially impractical for landscape scale application. While high-resolution aerial imagery enables detailed mapping, high acquisition and processing costs hinder scalability for broad regional surveillance \cite{pham2024, arasumani2021}.

A recent study by Pham et al. \cite{pham2024} using high-resolution imagery demonstrated success in classifying serrated tussock into multiple cover classes. However, despite advanced spectral and texture analysis techniques, areas with moderate serrated tussock cover levels (10–30\%) remain challenging to classify because its spectral response is influenced by both serrated tussock and co-occurring native grasses within the same pixel, leading to spectral mixing and reduced class separability. These moderate infestations often lack distinct phenological or textural cues, making them easily confused with surrounding vegetation. 

A key distinguishing feature of serrated tussock is its phenology, marked by seasonal changes in appearance and spectral response. The plant appears bright green in spring and summer, often bleaches in winter due to frost, and develops purple flower heads in late spring and summer \cite{osmond2008serrated, sheffield_sorting_2025}. These phenological shifts influence reflectance patterns captured by multispectral sensors, making temporal monitoring essential for improving classification accuracy \cite{McGowen2001b}.

The growing availability of satellite sensors such as Sentinel-2, which offer high spectral resolution and a five-day revisit frequency, presents a promising opportunity to develop scalable, satellite-based monitoring solutions. A recent study by Sheffield et al. demonstrated the use of multiple seasonal Sentinel-2 imagery to differentiate serrated tussock from native grasses by leveraging phenological variability across seasons \cite{sheffield_sorting_2025}. Building on this foundation, the present study evaluates a suite of models that incorporate seasonal features along with a range of vegetation indices that have been previously validated in general vegetation classification. Although not originally designed for serrated tussock detection, these indices are assessed here for their potential to enhance classification accuracy, particularly for the challenging medium cover class, where spectral ambiguity remains a significant obstacle.

The core motivation of this research is to determine whether the combined use of spectral diversity and multi-seasonal temporal features from Sentinel-2 imagery can compensate for lower spatial resolution and achieve classification accuracy comparable to or better than high-resolution aerial imagery. We hypothesize that temporal phenological dynamics and band-specific responses across seasons can be harnessed to distinguish even subtle class differences that are otherwise inseparable in single-date imagery. This study makes the following contributions:

\begin{enumerate}
    \item We have presented a systematic framework using Sentinel-2 imagery to classify serrated tussock across multiple cover classes by integrating spectral bands, texture measures, and temporal vegetation indices.
    \item We have tested a suite of models incorporating both phenological timing (seasonal composites: Autumn, Winter, Spring and Summer) and domain-validated vegetation indices.
    \item We have compared our classification results against the best performing aerial image-based method from Pham et al. \cite{pham2024}.
    \item We have also compared our classification results against the best phenological model from Sheffield et al. \cite{sheffield_sorting_2025}.
\end{enumerate}

\section{MATERIALS AND METHODS}

\subsection{Study Area}
\begin{figure}[!t]
    \centering
    \includegraphics[width=0.75\linewidth]{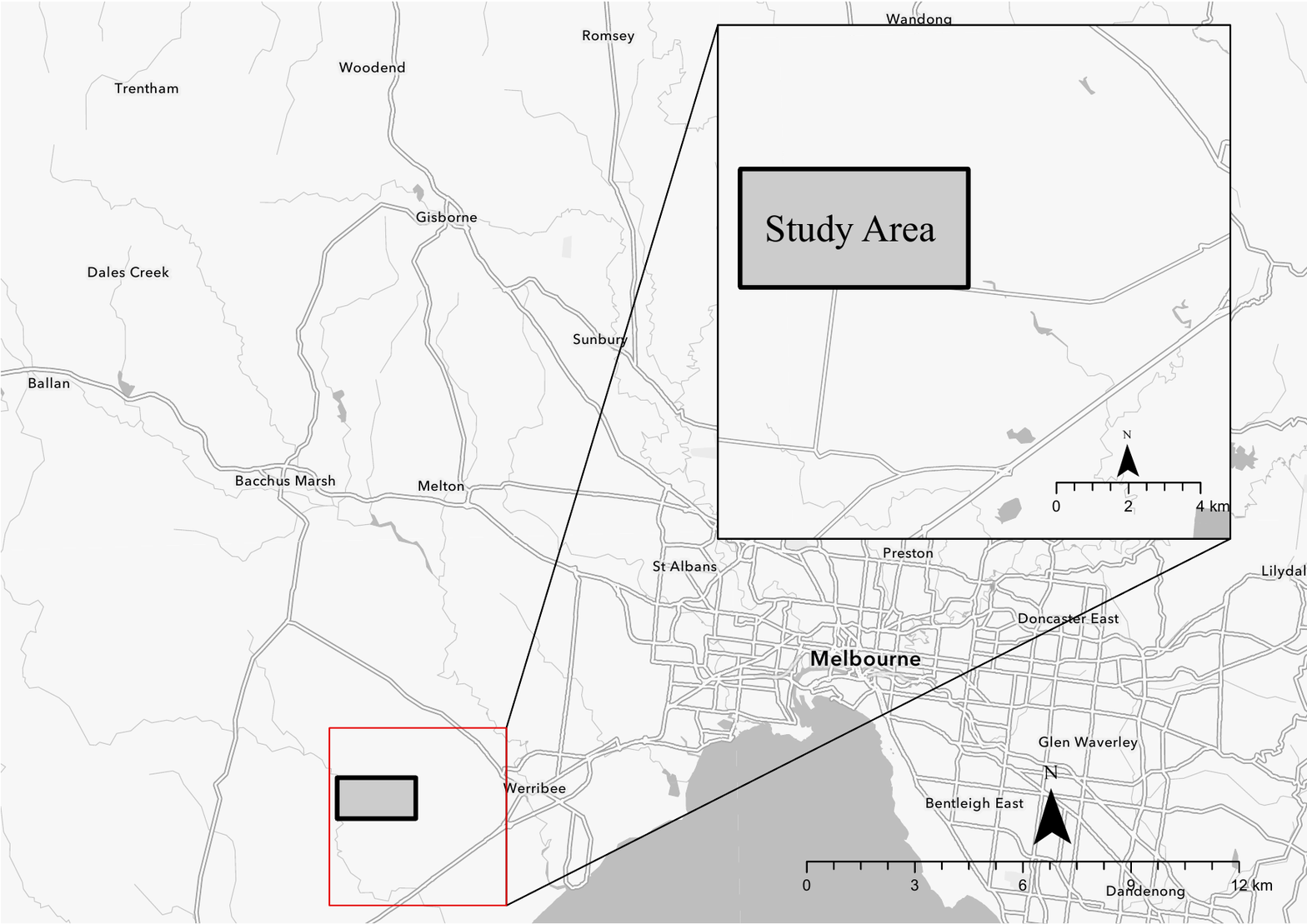}
    \caption{The study site, located in western Greater Melbourne, Victoria used to investigate the serrated tussock classification. }
    \label{fig:wgr}
\end{figure}

The study was conducted in the Western Greater Melbourne, covering approximately 1,200 hectares. The site was chosen on the presence of serrated tussock at varying cover levels, enabling a comprehensive analysis across an infestation gradient. The same location and ground truth data were used both in Sentinel-2 and aerial comparisons, allowing for direct comparison of results (Fig.~\ref{fig:wgr}).

\subsection{Datasets}
\subsubsection{Satellite data}
Sentinel-2 Level-2A (Bottom-of-Atmosphere) products were used from March 2021 to February 2023 \cite{esa2015sentinel2}. Table~\ref{tab:sentinel2_aerial_nm} shows the Sentinel-2 band details along with aerial imagery details for comparison \cite{pham2024}. Bands uninformative for vegetation (B1, B9, and B10) were excluded \cite{esa2015sentinel2}, and all 20m bands were resampled to 10m resolution using nearest neighbor (NN) sampling to ensure consistency across all inputs. NN was used to preserve original reflectance values, which makes it well-suited for classification tasks involving discrete vegetation classes \cite{google2024resample}. 
\begin{table}[htbp]
    \centering
    \caption{Comparison of Sentinel-2 Bands and RGB Aerial Imagery Used in \cite{pham2024}, with Wavelengths in 
    Nanometer (nm); CW = Central Wavelength, Res = Spatial Resolution (m).}
    \label{tab:sentinel2_aerial_nm}
    \renewcommand{\arraystretch}{1.2}
    \begin{tabular}{|l|c|c|l|c|c|}
        \hline
        \multicolumn{3}{|c|}{\textbf{Sentinel-2 MSI}} & \multicolumn{3}{c|}{\textbf{Aerial RGB Imagery }}\\
        \rowcolor{gray!20}
        \textbf{Channel (Band)} & \textbf{CW} & \textbf{Res} & \textbf{Channel} & \textbf{CW} & \textbf{Res} \\
        \hline
        Coastal Aerosol(B1)         & 443  & 60  & --       & --  & --    \\
        Blue(B2)                    & 490  & 10  & Blue     & 430 & 0.10  \\
        Green(B3)                   & 560  & 10  & Green    & 530 & 0.10  \\
        Red(B4)                     & 665  & 10  & Red      & 620 & 0.10  \\
        Red Edge 1(B5)              & 705  & 20  & --       & --  & --    \\
        Red Edge 2(B6)              & 740  & 20  & --       & --  & --    \\
        Red Edge 3(B7)              & 783  & 20  & --       & --  & --    \\
        NIR(B8)                     & 842  & 10  & --       & --  & --    \\
        Red Edge 4(B8A)             & 865  & 20  & --       & --  & --    \\
        Water Vapour(B9)           & 945  & 60  & --       & --  & --    \\
        SWIR-Cirrus(B10)         & 1,375 & 60  & --       & --  & --    \\
        SWIR 1(B11)                & 1,610 & 20  & --       & --  & --    \\
        SWIR 2(B12)                & 2,190 & 20  & --       & --  & --    \\
        \hline
    \end{tabular}
\end{table}
Nine vegetation indices were computed per season (ID1- NDVI \cite{ndviref}, ID2- NDWI \cite{ndwiref}, ID3- EVI \cite{eviref}, ID4- SAVI \cite{saviref}, ID5- IRECI \cite{ireciref}, ID6- NDVI std dev \cite{geeNeighborhoodReducer}, ID-7- TDVI \cite{tdviref}, ID8- NLI \cite{nliref}, ID9- MNLI \cite{mnliref}). Indices ID1-ID6 were adopted from Sheffield et al. \cite{sheffield_sorting_2025}, while ID7-ID9 were selected based on their success in other vegetation studies \cite{noauthor_broadband_nodate}. 

\subsubsection{Ground data}
Ground data for the study site were collected by the Victorian Department of Energy, Environment and Climate Action and Wyndham City Council. The assessment involved a systematic grid of 10m radius plots, spaced 40m apart both horizontally and vertically. Assessed plots were aligned with Sentinel-2 pixel footprints, made the data readily applicable to the Sentinel-2 analysis. The surveys were done in spring (September to November) and early summer (December) in 2021 and 2022 (spring and early summer). For the purpose of seasonal alignment with Sentinel-2 imagery, the year was divided into four standard meteorological seasons: Autumn (March–May), Winter (June–August), Spring (September–November), and Summer (December–February). 

Sentinel-2 imagery corresponding to the time span of the ground data collection was extracted, and seasonal composites were generated by calculating the median of cloud-free observations between September and December each year using Google Earth Engine \cite{gee2017}. This approach ensured that the spectral information used for classification closely represented the actual vegetation conditions observed during field surveys. 

A total of 6,879 plots were assessed across two years. The serrated tussock cover was classified into four cover classes: \textbf{None:} 0\% (660 plots), \textbf{Low:} 1-10\% (2,373 plots), \textbf{Medium:} 11-30\% (1,871 plots) and  \textbf{High:} $>$30\% (1,974 plots). These class divisions are used in the serrated tussock National Best Practice management guidelines \cite{osmond2008serrated}. 

\subsection{Feature Sets for Different Classification Models}
Table~\ref{tab:feature_set} presents the feature sets considered for elevent Sentinel-2 imagery based serrated tussock classification models. M17 to M76 replicate the baseline feature engineering strategy in \cite{pham2024}, which relied on RGB bands and Grey-level Co-occurence Matrix (GLCM) based texture features. Specifically; contrast, dissimilarity, homogeneity, energy, correlation, and angular second moment were extracted per band to capture spatial texture information \cite{pham2024, haralick}. 

\begin{table}[!t]
\centering
\caption{Summary of Features Used, Feature Counts, Number of Principal Components (PC) for the Sentinel-2 Imagery Models.}
\renewcommand{\arraystretch}{1.4}
\begin{tabular}{|c|p{5cm}|c|c|}
\hline
\rowcolor{gray!20}
\textbf{Model} & \textbf{Features Used} & \textbf{Count} & \textbf{PC}  \\ \hline
\textbf{M17} & B2, B3, B4, 14 spectral indices \cite{pham2024} & 17 & 4 \\ \hline
\textbf{M24} & GLCM (Six textures per band – B2, B3, B4, Grey) \cite{pham2024} & 24 & 13 \\ \hline
\textbf{M41} & Combination of M17 and M24 & 41 & 17 \\ \hline
\textbf{M10} & B2–B8, B8A, B11, B12 & 10 & 9 \\ \hline
\textbf{M66} & GLCM (Six textures per band – bands in M10, Grey) & 66 & 30 \\ \hline
\textbf{M76} & Combination of M10 and M66 & 76 & 9 \\ \hline
\textbf{M40} & B2–B8, B8A, B11, B12 (All season) & 40 & 40 \\ \hline
\textbf{M19} & B2–B8, B8A, B11, B12 + ID(1-9) (Field survey period) & 19 & 11\\ \hline
\textbf{M76*} & B2–B8, B8A, B11, B12 + ID(1-9) (All season) & 76 & 46  \\ \hline
\textbf{M64} & B2–B8, B8A, B11, B12 + ID(1-6) (All season) \cite{sheffield_sorting_2025} & 64 & 46  \\ \hline
\textbf{M20} & B2–B8, B8A, B11, B12 (Spring, Summer) & 20 & 20  \\ \hline
\end{tabular}
\label{tab:feature_set}
\end{table}

In contrast, M40 to M20 introduce novel contributions by leveraging the temporal richness and spectral diversity of Sentinel-2 imagery. These models incorporate multi-seasonal composites to exploit phenological variations, and integrate vegetation indices. M19 incorporating the same vegetation indices (ID1–ID9), but restricted to the field survey period only, allowing for direct comparison with multi-seasonal models.

\subsection{Classification Scenarios}

To reduce dimensionality and computation time, Principal Component Analysis (PCA) was applied prior to classification. The training features were scaled using StandardScaler in Python Programming Language with the scikit-learn 1.3.0 machine learning library \cite{scikit-learn}. The number of components was selected to preserve 99.9\% of variance \cite{KHERIF2020209}.

Each model was trained using a Random Forest classifier with 300 trees. For each model development, 80\% of the rapid plots were randomly selected for training (5,503 plots), while the remaining 20\% were withheld and used for validation (1,376 plots). A fixed random seed was used to ensure consistent sampling across model runs. The accuracy metrics used are provided in Table~\ref{tab:metrics} along with the equations \cite{kappa, assessment2}. 

\begin{table}[t]
\centering
\caption{Accuracy Metrics and Their Equations (TP = True Positive, TN = True Negative, FP = False Positive, FN = False Negative).}
\renewcommand{\arraystretch}{1.5}
\begin{tabular}{|l|c|}

\hline
\rowcolor{gray!20}
\textbf{Accuracy Metric} & \textbf{Equation} \\
\hline
Overall Accuracy (OA) & 
$\frac{TP + TN}{TP + FP + TN + FN}$
 \\ 
\hline
Expected Accuracy (EA) & 
$\frac{(TP + FP)(TP + FN) + (FN + TN)(FP + TN)}{(TP + TN + FN + FP)^2}$
 \\
\hline
Overall Kappa (OK) & 
$\frac{OA - EA}{1 - EA}
$ \\
\hline
F1-score (F1) & 
$\frac{2TP}{2TP + FP + FN}
$ \\
\hline
Precision (P) & 
$
\frac{TP}{TP + FP}
$ \\
\hline
Recall (R) & 
$
\frac{TP}{TP + FN}
$ \\
\hline
\end{tabular}
\label{tab:metrics}
\end{table}

\begin{figure}[htbp]
    \centering
    \includegraphics[width=\linewidth]{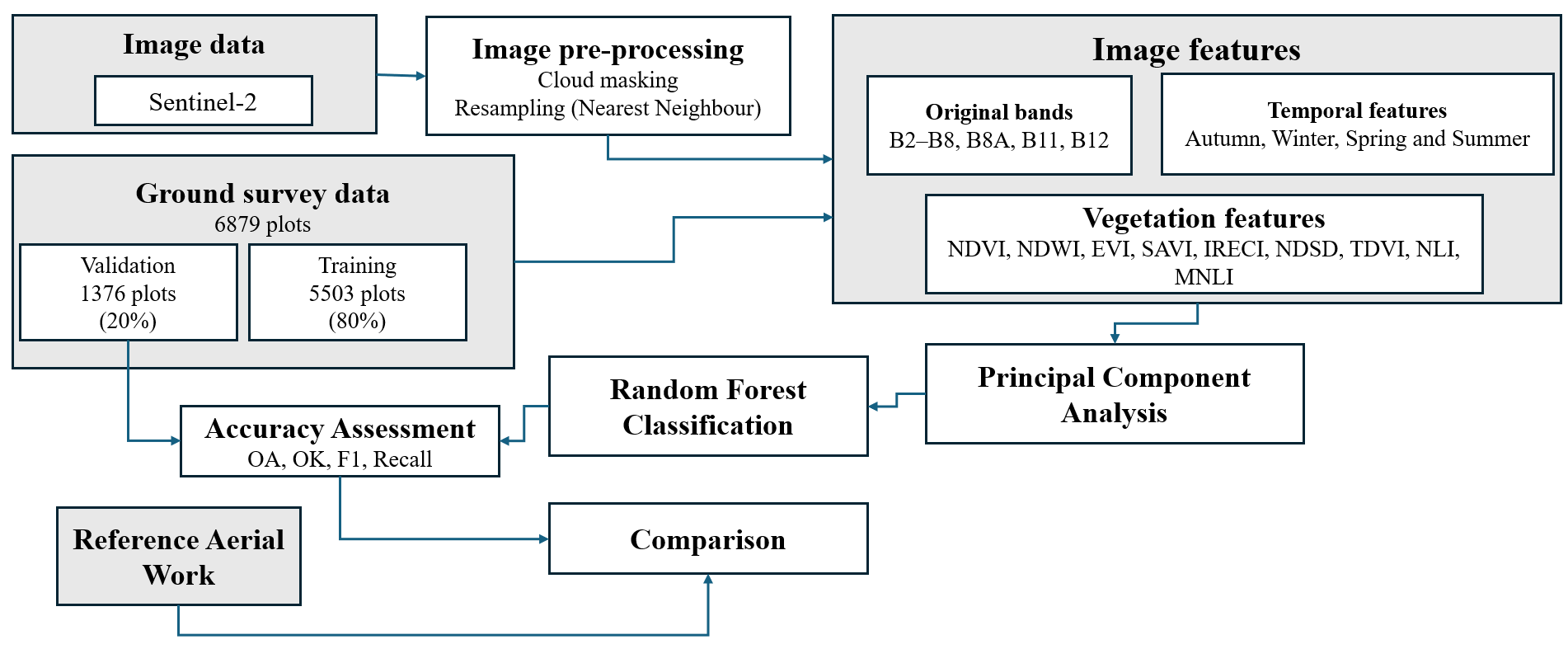}
    \caption{Workflow for serrated tussock classification using Sentinel-2.}
    \label{workflow}
\end{figure}

Fig.~\ref{workflow} depicts the workflow summary for serrated tussock classification using Sentinel-2. 

\section{Results and Discussion}
\subsection{Overall Performance Evaluation}
\begin{table*}[t]
\centering
\caption{Overall Accuracy (OA) and Overall Kappa (OK) Scores for All Models.}
\renewcommand{\arraystretch}{1.3}
\begin{tabular}{|c|c|c|c|c|c|c|c|c|c|c|c|c|}
\hline
 &M17 &	M24	&M41&	M10	&M66	&M76&	M40&	M76*&	M64&	M20&	M19&	Base result \cite{pham2024}\\
\hline
OA	&44\%	& 34\%&	50\%	&60\%	&36\%&	47\%	&67\%	&\textbf{68}\%	&66\%	&67\%	&61\%	&67\% \\
\hline
OK	&0.22&	0.04&	0.28&	0.43	&0.07&	0.23&	0.54&	\textbf{0.55}	&0.52	&0.54	&0.45&	0.52\\
\hline

\end{tabular}
\label{tab:overallAcc}
\end{table*}
Table~\ref{tab:overallAcc} summarizes the performance of all 11 models using Overall Accuracy (OA) and Overall Kappa (OK). Table~\ref{tab:result} summarizes some other performance evaluation metrics for selected models (M10, M40, M19, M76*, M64, M20). The results varied significantly depending on the feature set considered. OA ranged from 34\% to 68\%, and OK from 0.04 to 0.55. These differences reveal how sensitive model performance is to feature selection.
\begin{table*}[t]
\centering
\caption{Evaluation Metrics---F1-Score (F1), Precision (P), and Recall (R)---for All Cover Classes for Selected Models.}
\renewcommand{\arraystretch}{1.3}
\begin{tabular}{|c|ccc|ccc|ccc|ccc|}
\hline
\textbf{Model} 
& \multicolumn{3}{c|}{\textbf{None}} 
& \multicolumn{3}{c|}{\textbf{Low}} 
& \multicolumn{3}{c|}{\textbf{Medium}} 
& \multicolumn{3}{c|}{\textbf{High}} \\
\cline{2-13}
& \textbf{F1} & \textbf{P} & \textbf{R} 
& \textbf{F1} & \textbf{P} & \textbf{R} 
& \textbf{F1} & \textbf{P} & \textbf{R} 
& \textbf{F1} & \textbf{P} & \textbf{R} \\
\hline
M10 & 0.52 & 0.79 & 0.39 & 0.67 & 0.59 & 0.77 & 0.41 & 0.41 & 0.36 & 0.70 & 0.67 & 0.72 \\
M40 & 0.67 & 0.85 & 0.55 & 0.70 & 0.63 & 0.81 & 0.50 & 0.56 & 0.45 & 0.79 & 0.80 & 0.78 \\
M19 & 0.64 & 0.83 & 0.51 & 0.66 & 0.57 & 0.77 & 0.39 & 0.47 & 0.34 & 0.72 & 0.72 & 0.72 \\
M76* & 0.66 & \textbf{0.89} & 0.52 & 0.71 & 0.62 & \textbf{0.84} & \textbf{0.51} & \textbf{0.58} & 0.45 & \textbf{0.80} & \textbf{0.82} & 0.78 \\
M64 & 0.56 & 0.87 & 0.41 & 0.71 & 0.61 & \textbf{0.84} & 0.48 & \textbf{0.58} & 0.41 & 0.79 & 0.78 & \textbf{0.80} \\
M20 & \textbf{0.69} & \textbf{0.89} & \textbf{0.57} & 0.70 & 0.63 & 0.80 & 0.50 & 0.56 & 0.45 & 0.78 & 0.78 & 0.77 \\
Base Aerial Imagery Model \cite{pham2024} & 0.57 & 0.80 & 0.44 & \textbf{0.72} & \textbf{0.65} & 0.81 & \textbf{0.51} & 0.56 & \textbf{0.47} & 0.77 & 0.78 & 0.77 \\
\hline
\end{tabular}
\label{tab:result}
\end{table*}

Models such as \textbf{M24} and \textbf{M66}, which relied solely on GLCM-based texture features, delivered the weakest performance (OA = 34\% and 36\%, OK = 0.04 and 0.07, respectively). This confirms that texture features alone are inadequate in highly mixed landscapes where spectral differences among classes are subtle. 

In contrast, \textbf{M76*, M40, M20, and M64} consistently performed well across all metrics. \textbf{M76*} recorded the best results (OA = 68\%, OK = 0.55), followed closely by \textbf{M40} and \textbf{M20} (OA = 67\%, OK = 0.54) and \textbf{M64} (OA = 66\%, OK = 0.52). These models incorporated both \textbf{multi-seasonal composites} and \textbf{vegetation indices}, enabling them to leverage phenological variation and spectral richness effectively (Table~\ref{tab:feature_set}).

\subsection{Class-wise F1-Score and Recall Performance Insights}

\textbf{M20} achieved the highest F1 and recall for the “None” class (0.69 and 0.57), while \textbf{M76*} outperformed all others in the “Medium” (F1 = 0.51, Recall = 0.48) and “High” (F1 = 0.80, Recall = 0.78) classes. For the “Low” class, \textbf{M64} and \textbf{M76*} both achieved the top F1 (0.84), with M76* also recording the highest recall (0.84). Models enriched with seasonal inputs—\textbf{M40, M64, M76*, and M20}—consistently delivered superior performance across all classes. The lower recall for “Medium” cover is expected due to class ambiguity, aligning with findings from Pham et al. and Sheffield et al. \cite{pham2024, sheffield_sorting_2025}.

\subsection{Comparison of Omni- vs Multi-Seasonal Feature Sets}

\begin{itemize}
    \item \textbf{M19 vs. M76*}: Both models used the same 10-band feature set and all vegetation indices (ID1-ID9), but \textbf{M19} was limited to the ground survey period while \textbf{M76*} included all four seasonal composites. \textbf{M76*} demonstrated consistent improvements across all metrics: OA rose from 61\% to 68\%, OK from 0.45 to 0.55, F1 for Medium from 0.39 to 0.51, and Recall for Low from 0.57 to 0.84.
    \item \textbf{M10 vs. M40}: \textbf{M10} used raw Sentinel-2 bands from a single season; \textbf{M40} extended these over four seasons (Autumn, Winter, Spring and Summer). The seasonal expansion improved OA from 60\% to 67\% and enhanced class-wise performance across the board.
    \item \textbf{M20 vs. M40}: Both models shared the same input features (Sentinel-2 bands only), but \textbf{M20} was limited to two seasons (Spring and Summer), while \textbf{M40} utilized all four. \textbf{M40} achieved slightly higher OA (67\%) compared to \textbf{M20} (67\%) but with better F1 in the Medium class (0.50 vs. 0.45) and improved recall in None and Low classes. 
\end{itemize}

\subsection{Comparison Against Best Aerial Imagery Model in \cite{pham2024}}

Of the eleven models evaluated, Model \textbf{M76*} achieved the highest performance, with an OA of 68\% and an OK of 0.55, surpassing the best reference result from Pham et al. \cite{pham2024} (OA: 67\%, OK: 0.52). These results validate that Sentinel-2 imagery, despite its coarser spatial resolution, with temporal and spectral richness, can compensate for spatial resolution limitations, offering a cost-effective and scalable alternative to high-resolution aerial imagery. However, it is noteworthy that the “Medium” cover class (10–30\%) continues to pose a challenge across models, primarily due to spectral confusion between serrated tussock and co-occurring native grass species. 

\subsection{Comparison Against Best Phenology Model in \cite{sheffield_sorting_2025}}
Building upon the work of Sheffield et al. \cite{sheffield_sorting_2025}, which utilized a single year (2021) of Sentinel-2 imagery and six vegetation indices across a broader area (9,211 plots), the current study expands the analysis. Using two years of data (2021 and 2022) covering 6,879 plots, this study incorporates three additional vegetation indices: TDVI, NLI, and MNLI. Notably, \textbf{NLI and MNLI} model nonlinear reflectance interactions, which are common in dense vegetation due to light scattering. These indices enhanced performance in high-cover scenarios. Compared to Sheffield’s F1-scores (Medium: 0.49, High: 0.74), Model \textbf{M76*} achieved 0.51 and 0.80, respectively.

\section{Conclusion}

This study demonstrates that multi-temporal Sentinel-2 imagery, when enriched with vegetation indices and seasonal features, can rival and even surpass high-resolution aerial imagery in the classification of serrated tussock in heterogeneous grassland environments. While single-season and texture-based models struggled to accurately classify the medium cover class, however, models that incorporating full seasonal information such as \textbf{M76*} and \textbf{M40} demonstrated significantly improved classification performance, particularly in distinguishing low to medium infestations.
The best-performing model (\textbf{M76*}) achieved an OA of 68\% and an OK of 0.55, exceeding the benchmark established by aerial imagery. Moreover, the inclusion of three new vegetation indices enhanced classification accuracy.
While challenges persist in classifying medium cover due to spectral overlap with native grasses, this study highlights the potential of satellite-based monitoring as a cost-effective and scalable alternative management of invasive species. This work is currently limited to serrated tussock. Further investigation is needed to assess the approach's applicability to other tussock-forming grass species, supporting the development of broader, scalable monitoring frameworks for invasive grass.

\section{Acknowledgement}

This work was funded by the Victorian Department of Energy, Environment and Climate Action. The authors gratefully acknowledge the contributions of staff from TREC Land Services, Themeda Ecology, Practical Ecology, and ABZECO for their assistance in field data collection.

\bibliographystyle{IEEEtranN}
\bibliography{references}
\end{document}